\documentclass[10pt, conference, letterpaper]{ IEEEtran}
\IEEEoverridecommandlockouts
\usepackage[caption=false,font=normalsize,labelfont=sf,textfont=sf]{subfig}
\usepackage{cite}
\usepackage{amsmath,amssymb,amsfonts}
\usepackage{algorithmic}
\usepackage{graphicx}
\usepackage{textcomp}
\usepackage{xcolor}
\usepackage[algo2e,linesnumbered,lined,boxed,commentsnumbered,ruled]{algorithm2e}
\usepackage{makecell}
\usepackage{multirow}
\usepackage{color}
\def\BibTeX{{\rm B\kern-.05em{\sc i\kern-.025em b}\kern-.08em
    T\kern-.1667em\lower.7ex\hbox{E}\kern-.125emX}}
\begin{document}

%\title{SplitMe: Communication and Computation Efficient Split Federated Learning in O-RAN via Mutual Learning with Inverse Server-Side Model\\
% {\footnotesize \textsuperscript{*}Note: Sub-titles are not captured for https://ieeexplore.ieee.org  and
% should not be used}
%}

\title{Communication and Computation Efficient Split Federated Learning in O-RAN}

% \author{
% \IEEEauthorblockN{10 pages}
% }
\author{
\IEEEauthorblockN{Shunxian Gu\textsuperscript{*},
Chaoqun You\textsuperscript{*}\thanks{Chaoqun You is the corresponding author.},
Bangbang Ren\textsuperscript{\dag},
Deke Guo\textsuperscript{\ddag}}
\IEEEauthorblockA{\textit{\textsuperscript{*}Fudan University, Shanghai, China}\\
\textit{\textsuperscript{\dag}National University of Defense Technology, Changsha, China}\\
\textit{\textsuperscript{\ddag}Sun Yat-sen University, Guangzhou, China}}
}
\maketitle

\begin{abstract}
The hierarchical architecture of Open Radio Access Network (O-RAN) has enabled a new Federated Learning (FL) paradigm that trains models using data from non- and near-real-time (near-RT) Radio Intelligent Controllers (RICs). However, the ever-increasing model size leads to longer training time, jeopardizing the deadline requirements for both non-RT and near-RT RICs. To address this issue, split federated learning (SFL) offers an approach by offloading partial model layers from near-RT-RIC to high-performance non-RT-RIC. Nonetheless, its deployment presents two challenges: (i) Frequent data/gradient transfers between near-RT-RIC and non-RT-RIC in SFL incur significant communication cost in O-RAN. (ii) Proper allocation of computational and communication resources in O-RAN is vital to satisfying the deadline and affects SFL convergence. Therefore, we propose SplitMe, an SFL framework that exploits mutual learning to alternately and independently train the client-side model and the inverse server-side model between communication exchanges. Layer-wise analytic reconstruction then recovers the forward server-side model for inference. Then, we solve a joint optimization problem for  SplitMe to minimize overall resource costs with deadline-aware selection of near-RT-RICs and adaptive local updates. Our numerical results demonstrate that SplitMe remarkably outperforms FL frameworks like SFL, FedAvg and O-RANFed regarding costs and convergence.
\end{abstract}


\section{Introduction}

The Radio Access Network (RAN) has long become an essential component of 5G and 5G Beyond (NextG) networks, connecting end-user devices to the core network through radio connections \cite{agarwal2025open}. With the rapid increase in both the complexity and number of user demands, Open RAN (O-RAN) has been proposed as an open and disaggregated evolution of RAN, designed to promote interoperability, flexibility, vendor diversity, and intelligent network management \cite{you2025greenran,maxenti2024scalo}. This is done by disaggregating the processing layers from proprietary hardware to shared open clouds and installing RAN intelligent controllers (RICs) which utilize the operational and performance data to optimize the RAN resources \cite{sun2021harmonizing}.

%The O-RAN architecture has been the revolutionary step and published with two key concepts: (i) openness and (ii) in-built intelligence \cite{polese2023understanding}. It frees the network from vendor lock-in via disaggregated, virtualized and software-based components, increasing the resiliency
 % and reconfigurability of the RAN. Meanwhile, it also enables a smarter Service and Management Orchestration (SMO) by installing RAN intelligent controllers (RICs) which utilize the operational and performance data to optimize the RAN resources \cite{sun2021harmonizing}.
%  \textcolor{blue}{The O-RAN architecture, extensively detailed by the O-RAN Alliance, comprises six key components, namely: the SMO frameworks, non and near real-time RICs, O-Cloud, ORAN Central Unit (O-CU), ORAN Distributed Unit (O-DU), and ORAN Radio Unit
% (O-RU).}

 Benefit from these unique features, 5G use cases such as meeting the network slice Service Level Agreement (SLA) and Key Performance Indicator (KPI) monitoring can be well supported by the O-RAN architecture. All of these cases sometimes require fast and secure training of Machine Learning (ML) models. However, the traditional centralized training paradigm requires users to transfer their raw data to a central server, resulting in privacy leakage. Moreover, transferring the raw data to the server can result in a large bandwidth usage cost. Due to these reasons, the deployment of Federated Learning \cite{mcmahan2017communication}, which is an advanced ML tool for privacy protection and edge computing, has become an open problem in O-RAN \cite{lim2020federated}: \textit{how to minimize the communication resource usage cost and total training time while achieving a high-quality global model?} This joint optimization problem can be initially mathematically formulated as:
%  \begin{figure}[tbp]
% \centering
% \includegraphics[width=0.95\columnwidth]{overview_splitme.png}%
% \caption{Workflow of SFL and SplitMe.}
% \vspace{-0.5cm}
% \end{figure}
\begin{equation}
{\min}~{\rho}K_{\epsilon}^{co}\underbrace{(R^{co}+R^{cp})}_{\textbf{resource~cost}}+(1-\rho)K_{\epsilon}^{tr}\underbrace{\max\{T_m\}}_{\textbf{learning time}}
\label{eq1}
\end{equation}
where $\rho$ is the trade-off parameter. $K_{\epsilon}^{co}$ and $K_{\epsilon}^{tr}$ denote the number of global communication rounds and training rounds to achieve the $\epsilon$-accuracy respectively in FL. They are usually the same.
$R^{co}$ and $R^{cp}$ denote the cost of communication and computation resources. $T_m$ is the sum of the local computation time and the communication time to the central server per training round for the arbitrary $m$-th client. Many prior works \cite{singh2022joint,singh2024communication} have given solutions to the problem. However, with the surging data and model size, the local computation time of the clients with poor computational capability (e.g. resource-constrained edge server \cite{lin2024efficient,deng2022actions}) will become long and violate the stringent deadline of O-RAN control loops. To this end, Split Federated Learning (SFL) \cite{thapa2022splitfed,wu2023split} can serve as an effective approach to mitigate the computational burden by offloading partial workloads from the clients to a regional cloud server via layer-wise model partitioning.

However, integrating SFL into the O-RAN architecture is a challenging task, but a valuable research direction \cite{lee2024federated,huang2023validation}. \textbf{(i) Unaffordable communication resource usage cost.} In the vanilla SFL, the activation values of the split layer (called the smashed data) in feed-forward computing and its gradients in backward propagation are communicated (uploaded and downloaded) between the clients and the cloud server when each batch of data is processed. This batch-level transmission can bring a greater number of global communication rounds (i.e. $K_{\epsilon}^{co}$) than conventional FL approaches, which conduct the communication round after processing the whole local dataset, resulting in unaffordable communication resource usage cost. Furthermore, although some prior SFL frameworks \cite{han2021accelerating,he2020group,ayad2021improving,wu2024ecofed,zheng2023reducing,oh2025communication} (which are not primarily designed for the O-RAN architecture) have made contributions to reducing the communication cost, they suffer from additional computational overhead, a narrower application scope, and the risk of divergence (see section \ref{p2p1}), degrading the Grade of Service (GoS) in the O-RAN system. None of the previous works have simultaneously addressed such sufferings \textit{which motivate us to develop a new SFL framework to reduce the communication resource usage cost for the O-RAN architecture.} \textbf{(ii) Resource allocation to satisfy the stringent deadline of O-RAN control
loops.} In the novel FL frameworks for the O-RAN architecture, the control loop deadline is the maximum time consumption permitted in each interval of the global communication. Since SFL can bring extra communication for transferring intermediate feature or gradient matrix, how to allocate the bandwidth resources and conduct the selection of clients becomes critical for satisfying the control loop. However, current solutions have not considered the SFL setting yet and they assume a larger fraction of clients in each round saves the total time required to attain the desired model accuracy. They ignore the effect of the number of local updates on the computational resource cost and learning time. \textit{This motivates us to reformulate the joint optimization problem for our proposed SFL framework and propose a novel resource allocation solution to it while not violating the deadline of O-RAN control loops.}

\begin{table}[!tbp]
\renewcommand{\arraystretch}{1.0}
\setlength\tabcolsep{1.8pt}
\caption{Comparison with Existing Literature on FL and SFL.}
\begin{center}
\centering
\begin{tabular}{c|c|c|c|c|c}
%添加顶部横线 
\Xhline{1.5 pt}
%输入标题
% \cite{2-3}
% \cline{2-7}
% \Xhline{1.5 pt}
%添加标题和内容之间的横线
% \Xhline{0.5 pt}
\centering
% \cline{2-4}
\multirow{2}{*}{Methods}&\multirow{2}{*}{SFL}&\multirow{2}{*}{Participants}&Auxiliary&Use Pre-trained&Divergence\\
&&&Network&Model&Risk\\
\Xhline{0.5 pt}
\multirow{2}{*}{DPFL\cite{han2021accelerating}}&\multirow{2}{*}{Yes}&{Edge Devices,}&\multirow{2}{*}{Yes}&\multirow{2}{*}{No}&\multirow{2}{*}{No}\\
&&Central Server&&&\\
% \multirow{2}{*}{EcoFed\cite{wu2024ecofed}}&\multirow{2}{*}{Yes}&\multirow{2}{*}{Edge Devices,}&\multirow{2}{*}{No}&\multirow{2}{*}{Yes}&\multirow{2}{*}{No}\\
% &&Central Server&&&\\
\multirow{2}{*}{EcoFed\cite{wu2024ecofed}}&\multirow{2}{*}{Yes}&{Edge Devices,}&\multirow{2}{*}{No}&\multirow{2}{*}{Yes}&\multirow{2}{*}{No}\\
&&Central Server&&&\\
MCOFed\cite{singh2024communication}&No&O-RAN RICs&No&No&Yes\\
O-RANFed\cite{singh2022joint}&No&O-RAN RICs&No&No&No\\
SplitMe&Yes&O-RAN RICs&No&No&No\\
% $\textbf{z}_{C,m}^*$&the label of one data sample on the $m$-th client, which is\\& generated from the optimal inverse server-side model\\
% $p_{C,m}^t(\textbf{z})$& the probability distribution of $\textbf{z}_{C,m}^t$, which is time-varying\\
% $p_{C,m}^*(\textbf{z})$& the probability distribution of $\textbf{z}_{C,m}^*$\\
% $d_{C,m}^t$&$\int\|{p_{C,m}^t(\textbf{z})-p_{C,m}^*(\textbf{z})}\|d\textbf{z}$\\
% ${\nabla}{f_{C,m}(\textbf{w};\textbf{z})}$&the gradient from one data sample on the $m$-th client\\
% $q_m$&the probability of choosing the $m$-th client,$\sum_{m=1}^M{q_m}=1$\\
% % $f_C(\textbf{w}_{C})$&the global loss function of the client-side model\\
% % $F_C(\textbf{w}_{C})$&the global loss function of the client-side model when the\\& inverse server-side model is optimal\\
% $\tilde{\nabla}{F_{C,m}(\textbf{w}_{C})}$&the gradient computed on a data batch $\xi_m^t\subset\tilde{\mathcal{D}}_m$ using\\& the local loss function $F_{C,m}(\textbf{w}_{C})$\\
% % ${\nabla}{F_{C,m}(\textbf{w}_C)}$&$\mathbb{E}[\tilde{\nabla}{F_{C,m}(\textbf{w}_C)}]$\\
% $\nabla{F_C(\textbf{w}_{C})}$&${\frac{1}{K}\sum_{m\in\mathcal{A}_t}({\nabla}{F_{C,m}(\textbf{w}_{C})}}=\mathbb{E}[\tilde{\nabla}{F_{C,m}(\textbf{w}_C)}])$\\
\Xhline{1.5 pt}
\end{tabular}
\label{features}
\end{center}
\vspace{-0.6cm}
\end{table}
In this paper, we first propose an SFL framework, namely SplitMe, that achieves independent local training for both the clients (near-RT-RICs) and the cloud server (non-RT-RIC) to greatly reduce the transmission between them. Different from previous FL frameworks for the O-RAN architecture and SFL frameworks, specifically, we propose a novel mutual learning approach to alternately learn the client-side model and the inverse server-side model. For inference, SplitMe combines the client-side model with the forward server-side model recovered by layer-wise analytic reconstruction \cite{zhuang2025analytic}. Other characteristics of SplitMe are listed in the Table. \ref{features} and explained in section \ref{p2p1}. Then, we reformulate the joint optimization problem of the equation \ref{eq1} for our SplitMe and propose a resource allocation algorithm with deadline-aware selection of local trainers (i.e. clients) and adaptive local updates to solve the joint optimization problem. Our contributions can be summarized as follows:

\begin{itemize}
    \item We propose SplitMe, which achieves independent local training for both the near-RT-RICs and non-RT-RIC to greatly reduce the transmission between them in SFL and gives its convergence analysis. This is the first SFL framework for the O-RAN architecture to the best of our knowledge.
    \item We reformulate the joint optimization problem in the O-RAN architecture to the SFL setting to minimize communication resource usage cost and total training time under the constraints of limited bandwidth, selected local trainers and control loop deadline.
    \item We propose a solution to this problem by decomposing it into the subproblems of resource allocation and suboptimal local trainers selection. The subproblems are resolved alternately at the beginning of each global training round, achieving adaptive client selection and local updates.
    \item Our numerical results on the O-RAN COMMAG dataset indicate that SplitMe is both computation- and communication-efficient and it remarkably outperforms FL frameworks like SFL, FedAvg and O-RANFed in terms of costs and convergence.
\end{itemize}

\section{Related Works}\label{p2p1}
\textbf{Federated Learning (FL) in O-RAN}: Different from the classical FL paradigm that utilizes UEs and a base station \cite{dinh2020federated} or edge devices and a central server \cite{xu2020client} to train a global model, FL in O-RAN treats near-RT-RICs and a non-RT-RIC as the clients and the server to train the model. Typically, the near-RT-RICs can be edge servers with low computational capability (e.g. base stations and access points) while the non-RT-RIC can be a regional cloud server with high computational capability. Such a new paradigm requires that the time consumption of each training round should not exceed the stringent deadline of O-RAN control
loops, which is solved in O-RANFed\cite{singh2022joint}. To further improve the scalability of clients under the stringent deadline, MCORANFed \cite{singh2024communication} is proposed by compressing the model updates in each communication round. \textit{However, none of them consider the SFL setting where the near-RT-RICs are resource-constrained and the model and data size become bigger.}

\textbf{Split Federated Learning (SFL)}: SFL is considered a variant of FL to address the limitation of operating FL on resource-constrained clients by partitioning deep neural
network (DNN) workloads between the local clients and the cloud server. In SFL, each client transmits the smashed data to the regional cloud server and receives the gradients from the server for further backpropagation to update the client-side model. Such a way causes the challenge of communication overhead and many prior works have made efforts to address it. The first approach is to decrease the frequency of intermediate feature or gradient matrix transfer. In \cite{han2021accelerating,he2020group}, local loss is generated by an auxiliary network to train the client-side model instead of transferring and using the gradient from the server. However, the introduction of the auxiliary network can bring an extra computational burden, especially when the input dimensionality of the auxiliary network is high. The second approach employs a pre-trained autoencoder to reduce the dimensionality of the smashed data, where the encoder is inserted at the output of the client-side model, and the decoder is inserted at the input of the server-side model \cite{ayad2021improving,wu2024ecofed}. However, such an approach requires a pre-trained client-side model which is impractical for most ML tasks that need learning from scratch. The third approach is to use sparsification or quantization techniques to compress the intermediate feature and gradient matrices.  In \cite{zheng2023reducing}, the authors enhance the top-S technique by introducing randomness to improve its performance. However, the compression level is hard to control, thus resulting in the risk of model divergence. \textbf{To sum up, although these prior works can reduce communication cost, they suffer from additional computational overhead, a narrower application scope, and the risk of divergence}.

Since SpliMe does not require an auxiliary network and a pre-trained client-side model, and has a convergence guarantee (see section \ref{p3p3}), it escapes from the sufferings above.

\section{Design of SplitMe}
\subsection{System Model}
\begin{figure}[tbp]
\centering
\includegraphics[width=0.9\columnwidth]{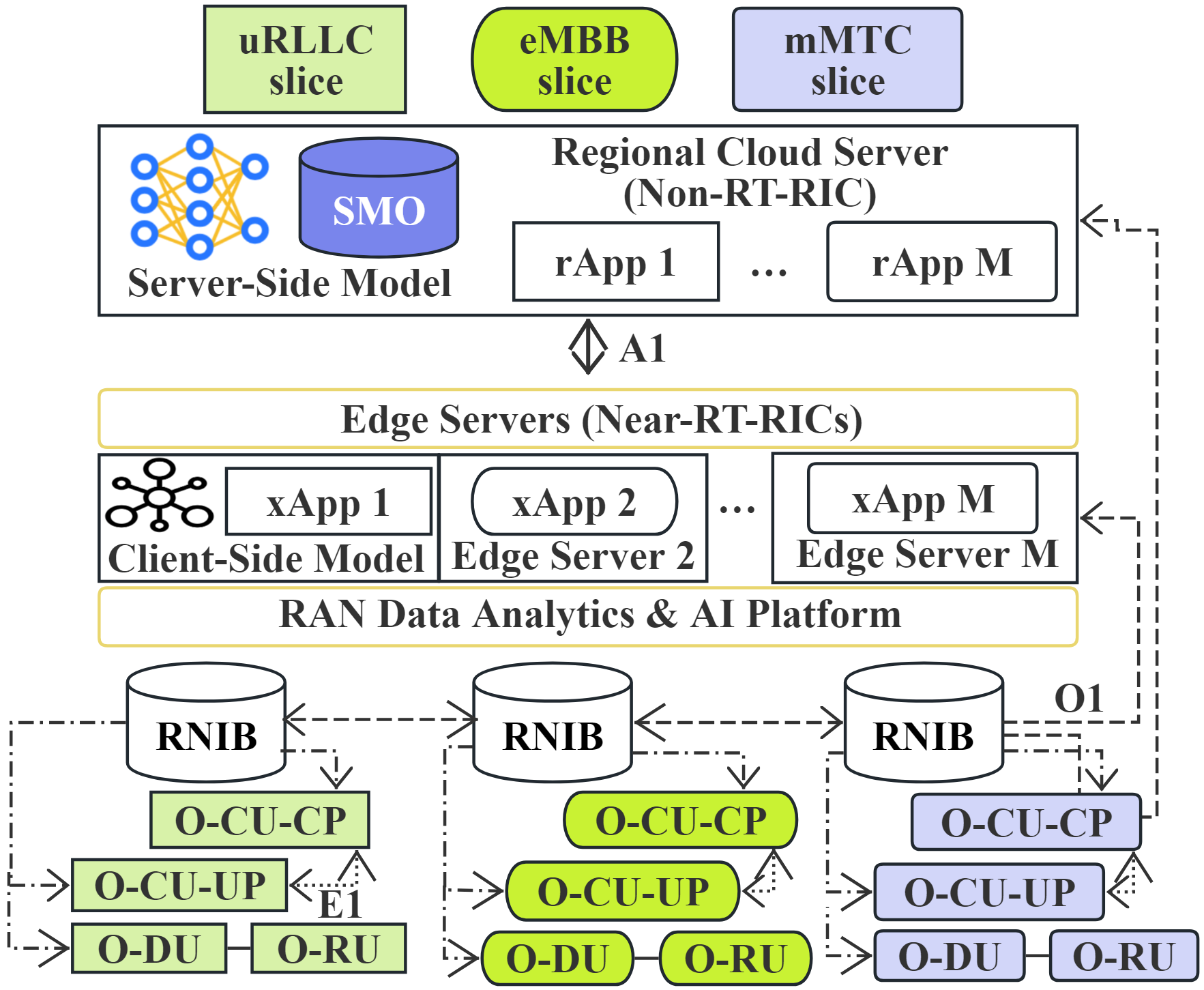}%
\caption{Overview of SplitMe system model.}
\label{system_model}
\vspace{-0.5cm}
\end{figure}

Similar to \cite{singh2022joint,singh2024communication}, we consider an O-RAN system with one regional cloud server (the server) and a set $\mathcal{M}$ of $M$ distributed edge servers (the clients) collaboratively training a model. As shown in Fig. \ref{system_model}, we introduce the essential components for training such an FL model as follows:

\textbf{Radio Network Information Base (RNIB) and E2 interface}: The E2 interface is a logical interface that lets near-RT-RIC collect performance measurements (PM) and operational data from O-DU, O-CU-UP and O-eNB \cite{polese2023understanding}. The collected data are stored in RNIB and shared by multi-vendor edge servers.

\textbf{O1 interface}: Each xApp on the near-RT-RIC collects the data for model training from one or more RNIBs via the O1 interface. The collected data are differentiated traffic PM for each RAN slice. We denote the collected data by $\mathcal{D}_m$ as the PM data of $m$-th near-RT-RIC instance and $\mathcal{D}_{i}\cap \mathcal{D}_{j}=\emptyset,\ \forall\ i\neq j$. Meanwhile, just as the traditional SFL setting, the labels of the collected data are transmitted to the rApp on the non-RT-RIC that corresponds to the xApp, through the O1 interface.

\textbf{xApp, rApp and A1 interface.}
Each xApp runs on a near-RT-RIC and trains a client-side model
$c(\cdot;\mathbf{w}_C)$. Its corresponding rApp runs on the
non-RT-RIC and trains an inverse server-side model
$s^{-1}(\cdot;\mathbf{w}_S)$ that maps labels to cut-layer targets.
After mutual learning, layer-wise analytic reconstruction yields the
parameters $\mathbf{w}_S^*$ of the forward server-side model
$s(\cdot;\mathbf{w}_S^*)$. The A1 interface carries training messages
between xApps and rApps. The rApps communicate with one another
through GLOO \cite{incubator2017gloo}. We refer to the participating
near-RT-RICs as clients and the non-RT-RIC as the server.

\subsection{Workflow of SplitMe}
Conventional SFL partitions a prediction model into a client-side
model $c$ and a forward server-side model $s$. Its output-space
objective can be written as
\begin{gather}
    \min_{\mathbf{w}_C,\mathbf{w}_S^{\mathrm{fwd}}}
    \frac{1}{M}\sum_{m\in\mathcal{M}}
        D_{\mathrm{out}}\big(
            s(c(\mathbf{X}_m;\mathbf{w}_C);\mathbf{w}_S^{\mathrm{fwd}}),
            \mathbf{Y}_m\big),\\
    (\mathbf{X}_m,\mathbf{Y}_m)=\mathcal{D}_m ,
    \label{eq2}
\end{gather}
where $\mathbf{X}_m$ contains the samples of xApp $m$ and
$\mathbf{Y}_m$ contains their ordered one-hot labels.
Here $\mathbf{w}_S^{\mathrm{fwd}}$ parameterizes the forward server-side
model in conventional SFL, and $D_{\mathrm{out}}$ measures prediction
error.
To train this model in SFL, each xApp uploads its smashed data and
waits for the cut-layer gradient for every mini-batch.

\textbf{Representation matching.}
SplitMe enables multiple local updates between communication exchanges
through the inverse server-side model $s^{-1}(\cdot;\mathbf{w}_S)$.
It maps each label to a target in the same $d_z$-dimensional space
as the client-side output. This reverse mapping motivates the surrogate
\begin{equation}
    \mathcal{L}_{\mathrm{match}}(\mathbf{w}_C,\mathbf{w}_S)
    =\frac{1}{M}\sum_{m\in\mathcal{M}}
      \varphi_m(\mathbf{w}_C,\mathbf{w}_S),
    \label{eq3}
\end{equation}
where $\varphi_m$ averages the following discrepancy over
sample--label pairs $(\mathbf{x},\mathbf{y})\in\mathcal{D}_m$:
$D(c(\mathbf{x};\mathbf{w}_C),s^{-1}(\mathbf{y};\mathbf{w}_S))$.
Here $D(\cdot,\cdot)$ is a nonnegative discrepancy measure satisfying
$D(\mathbf{A},\mathbf{B})=0$ if and only if $\mathbf{A}=\mathbf{B}$.
The notation $s^{-1}$ describes the direction of the learned mapping
from labels to representations.
Training alternates between two block subproblems with the other
block's targets held fixed:
\begin{equation}
\left\{
\begin{aligned}
    &\min_{\mathbf{w}_C}\ \mathcal{L}_{\mathrm{match}}(\mathbf{w}_C,\widetilde{\mathbf{w}}_S),\\
    &\min_{\mathbf{w}_S}\ \mathcal{L}_{\mathrm{match}}(\widetilde{\mathbf{w}}_C,\mathbf{w}_S).
\end{aligned}
\right.
\label{eq4}
\end{equation}
Here $\widetilde{\mathbf{w}}_C$ and $\widetilde{\mathbf{w}}_S$ denote the
parameters held fixed in their respective subproblems.
In the first subproblem, the inverse server-side model supplies targets
for the client-side model. In the second, each rApp uses smashed data
from the endpoint of its xApp's local training to supervise the
inverse server-side model. Thus both sides optimize the same
representation discrepancy through mutual learning.

\textbf{Initialization.}
The non-RT-RIC initializes the global models
$\bar{\mathbf{w}}_C^{(0)}$ and $\bar{\mathbf{w}}_S^{(0)}$.
Each xApp sends its ordered labels once to its corresponding rApp
to align them with later smashed-data uploads.
Let $k=1,\ldots,K_\epsilon$ index global rounds as in
Algorithm~\ref{al2}, and let $t$ index local SGD updates as in
Section~\ref{p3p3}. Round $k$ starts at local iteration $t_k$ and
runs for $E$ local steps per model, with $t_1=0$ and
$t_{k+1}=t_k+E$. Here $E$ is the local update budget for the
current round. Let $\mathcal{A}_k$ be the nonempty set of clients
selected for round $k$ and let $K=|\mathcal{A}_k|$.
For the convergence analysis, write $\mathcal{A}(t)=\mathcal{A}_k$
when $t_k\leq t<t_{k+1}$.

\textbf{Step 1 (Model and target download).}
The rApp generates one target
$\mathbf{r}_a^k=s^{-1}(\mathbf{e}_a;\bar{\mathbf{w}}_S^{(t_k)})$
for each class $a$ present at xApp $m$, where $\mathbf{e}_a$ is
its one-hot label. The xApp downloads $\bar{\mathbf{w}}_C^{(t_k)}$
and the class-wise targets once, then assigns each sample the
target of its class. It forms the local dataset
$\tilde{\mathcal{D}}_m=(\mathbf{X}_m,\widehat{\mathbf{Z}}_m^k)$
by storing the targets in the same sample order as $\mathbf{X}_m$.
For a given sample, its row of $\widehat{\mathbf{Z}}_m^k$ gives
the target $\mathbf{z}_{C,m}^t$ in Table~\ref{tab1} throughout
round $k$.

\textbf{Step 2 (Client-side model update).}
The xApp initializes
$\mathbf{w}_{C,m}^{(t_k)}=\bar{\mathbf{w}}_C^{(t_k)}$ and
performs $E$ mini-batch SGD updates. For $t_k\leq t<t_{k+1}$,
\begin{equation}
    \mathbf{w}_{C,m}^{(t+1)}=\mathbf{w}_{C,m}^{(t)}
    -\eta_C\,\tilde{\nabla}f_{C,m}(\mathbf{w}_{C,m}^{(t)}),
    \label{eq5}
\end{equation}
where $\eta_C$ is the client-side step size and
$\tilde{\nabla} f_{C,m}$ is evaluated on a mini-batch
$\xi_m^t\subset\tilde{\mathcal{D}}_m$. The loss $f_{C,m}$ measures
the discrepancy $D$ between the client output and the fixed
targets in $\widehat{\mathbf{Z}}_m^k$. The xApp uses the local
endpoint to compute the full smashed-data matrix in sample order,
$\mathbf{Z}_m^k=c(\mathbf{X}_m;\mathbf{w}_{C,m}^{(t_{k+1})})$,
and uploads it along with the local model parameters.

\textbf{Step 3 (Inverse server-side model update and aggregation).}
The rApp forms $\check{\mathcal{D}}_m=(\mathbf{Y}_m,\mathbf{Z}_m^k)$
using the stored labels. Starting from
$\mathbf{w}_{S,m}^{(t_k)}=\bar{\mathbf{w}}_S^{(t_k)}$, it takes $E$
local SGD steps. For $t_k\leq t<t_{k+1}$,
\begin{equation}
    \mathbf{w}_{S,m}^{(t+1)}=\mathbf{w}_{S,m}^{(t)}
    -\eta_S\,\tilde{\nabla}f_{S,m}(\mathbf{w}_{S,m}^{(t)}),
    \label{eq6}
\end{equation}
where $\eta_S$ is the inverse-model step size and
$\tilde{\nabla} f_{S,m}$ is evaluated on a mini-batch
$\Xi_m^t\subset\check{\mathcal{D}}_m$. The loss $f_{S,m}$ uses
the same discrepancy $D$ with $\mathbf{Z}_m^k$ fixed. The
non-RT-RIC averages the local endpoints of the two models separately
before starting the next round:
\[
    \bar{\mathbf{w}}_j^{(t_{k+1})}
    =\frac{1}{K}\sum_{m\in\mathcal{A}_k}
        \mathbf{w}_{j,m}^{(t_{k+1})},
    \qquad j\in\{C,S\}.
\]
The cut-layer traffic in a training round is one target downlink and
one smashed-data uplink per participating xApp, independent of how
many local mini-batch updates are taken between exchanges.

\begin{figure}[tbp]
\centering
\includegraphics[width=0.8\columnwidth]{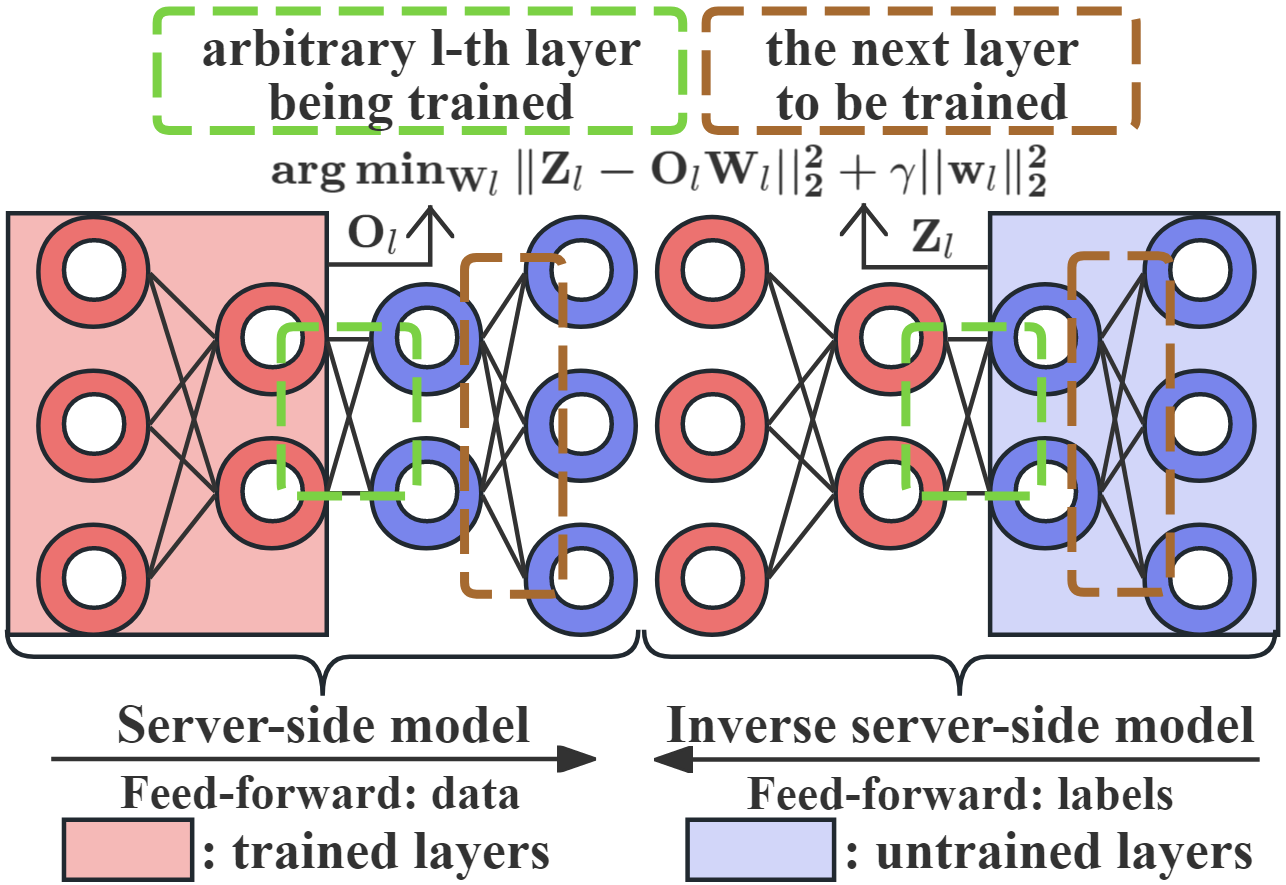}%
\caption{Layer-wise reconstruction of the forward server-side model.}
\label{zeroth}
\vspace{-0.5cm}
\end{figure}

\textbf{Step 4 (Final model acquisition).}
After the last round, SplitMe recovers the forward server-side
model $s$ by layer-wise analytic reconstruction
\cite{zhuang2025analytic,he2025afl,toh2025analytic}. The reconstruction
uses the participants in the final round,
$\mathcal{R}=\mathcal{A}_{K_\epsilon}$.
Let $T=t_{K_\epsilon+1}$ be the total number of local iterations.
Each selected xApp recomputes its smashed data
$c(\mathbf{X}_m;\bar{\mathbf{w}}_C^{(T)})$ in the original sample
order and uploads it once. The corresponding rApp feeds the ordered
one-hot labels $\mathbf{Y}_m$ through the aggregated inverse
server-side model $s^{-1}(\cdot;\bar{\mathbf{w}}_S^{(T)})$.
Let $\mathbf{H}_j^{(m)}$ denote the activated output after the
first $j$ inverse layers, counted from the label input. For a
forward server-side model with $L_s$ layers, the layer targets are
\[
    \mathbf{H}_0^{(m)}=\mathbf{Y}_m,\qquad
    \mathbf{Z}_\ell^{(m)}=\mathbf{H}_{L_s-\ell}^{(m)},
    \quad \ell=1,\ldots,L_s .
\]
The activated inverse output $\mathbf{Z}_\ell^{(m)}$ serves as the
expected pre-activation of forward layer $\ell$. In particular, the
last forward layer uses $\mathbf{Z}_{L_s}^{(m)}=\mathbf{Y}_m$.
Figure~\ref{zeroth} illustrates this reverse correspondence.

The non-RT-RIC fits the forward layers in order from $\ell=1$ to
$L_s$. Let $\mathbf{O}_\ell^{(m)}$ contain the input to layer
$\ell$ with a column of ones appended. The input to the first
layer is the refreshed smashed data. At each layer, for $\gamma>0$,
the non-RT-RIC fits the augmented weight and bias matrix
$\boldsymbol{\Theta}_\ell=[\mathbf{W}_\ell;\mathbf{b}_\ell^\top]$ by solving
\begin{equation}
    \min_{\boldsymbol{\Theta}_\ell}
    \sum_{m\in\mathcal{R}}
    \|\mathbf{O}_\ell^{(m)}\boldsymbol{\Theta}_\ell
      -\mathbf{Z}_\ell^{(m)}\|_F^2
      +\gamma\|\boldsymbol{\Theta}_\ell\|_F^2 .
    \label{eq7}
\end{equation}
Define the two sums
\[
\begin{aligned}
    \mathbf{A}_\ell
    &=\sum_{m\in\mathcal{R}}
      (\mathbf{O}_\ell^{(m)})^\top\mathbf{O}_\ell^{(m)},\\
    \mathbf{B}_\ell
    &=\sum_{m\in\mathcal{R}}
      (\mathbf{O}_\ell^{(m)})^\top\mathbf{Z}_\ell^{(m)} .
\end{aligned}
\]
The ridge solution $\boldsymbol{\Theta}_\ell^*$ satisfies
\begin{equation}
    (\mathbf{A}_\ell+\gamma\mathbf{I})\boldsymbol{\Theta}_\ell^*
    =\mathbf{B}_\ell .
    \label{eq8}
\end{equation}
The rApps can obtain the two sums by all-reduce. After solving
\eqref{eq8}, the non-RT-RIC applies the forward activation
$\sigma_\ell$ and appends a column of ones for the next layer:
\[
    \mathbf{O}_{\ell+1}^{(m)}
    =[\sigma_\ell(\mathbf{O}_\ell^{(m)}\boldsymbol{\Theta}_\ell^*),
      \mathbf{1}],\qquad \ell<L_s .
\]
The recovered weights and biases of all $L_s$ layers form
$\mathbf{w}_S^*$, which is combined with
$\bar{\mathbf{w}}_C^{(T)}$ to produce $s(c(\cdot))$ for inference.

In the convergence analysis, $f_{C,m}$ and $f_{S,m}$ use the
current targets. The reference losses $F_{C,m}$ and $F_{S,m}$ use
targets from the optimal counterpart model. The global reference
loss for block $j\in\{C,S\}$ is
$F_j(\mathbf{w}_j)=M^{-1}\sum_{m\in\mathcal{M}}F_{j,m}(\mathbf{w}_j)$.
Table~\ref{tab1}
denotes the current and reference client-side targets by
$\mathbf{z}_{C,m}^t$ and $\mathbf{z}_{C,m}^*$, respectively.
It uses $q_m$ for client selection in the analysis. The model
average above is uniform over the clients selected in the round.

\subsection{Convergence Analysis}\label{p3p3}
For clarity, we mainly focus on the convergence analysis of the client-side model while Corollary 3 treats the inverse server-side model. We list some notations required in the convergence analysis in the Table \ref{tab1}. The theory of SplitMe is based on the following assumptions, theorems and corollaries.
\begin{table}[!tbp]
\renewcommand{\arraystretch}{1.0}
\setlength\tabcolsep{1.8pt}
\caption{Table of partial notations.}
\begin{center}
\centering
\begin{tabular}{c|l}
%添加顶部横线 
\Xhline{1.5 pt}
%输入标题
% \cite{2-3}
% \cline{2-7}
% \Xhline{1.5 pt}
%添加标题和内容之间的横线
% \Xhline{0.5 pt}
\centering
% \cline{2-4}
Notations&Description\\
\Xhline{0.5 pt}
$\mathbf{z}_{C,m}^t$&Cut-layer target used by client $m$ at local step $t$,\\
& generated by the inverse server-side model\\
$\mathbf{z}_{C,m}^*$&Reference cut-layer target from the optimal\\
& inverse server-side model\\
$p_{C,m}^t(\mathbf{z})$&Distribution of the current target $\mathbf{z}_{C,m}^t$\\
$p_{C,m}^*(\mathbf{z})$&Distribution of the reference target $\mathbf{z}_{C,m}^*$\\
$d_{C,m}^t$&$\int \|p_{C,m}^t(\mathbf{z})-p_{C,m}^*(\mathbf{z})\|\,d\mathbf{z}$\\
$d_{S,m}^t$&Analogous distance for the rApp's smashed-data targets\\
$\nabla f_{C,m}(\mathbf{w};\mathbf{z})$&Client-side gradient for a sample with target $\mathbf{z}$\\
$\mathcal{A}(t)$, $K$&Active clients at local step $t$ and their count\\
$q_m$&Client-selection weight in the analysis, $\sum_{m=1}^M q_m=1$\\
$\tilde{\nabla} f_{C,m}(\mathbf{w}_C)$&Mini-batch gradient of the current client-side loss\\
$\tilde{\nabla} F_{C,m}(\mathbf{w}_C)$&Mini-batch gradient of the reference client-side loss\\
$\nabla F_C(\mathbf{w}_C)$&Gradient of the global client-side reference loss\\
\Xhline{1.5 pt}
\end{tabular}
\label{tab1}
\end{center}
\vspace{-0.6cm}
\end{table}

\textbf{Assumption 1.} \textit{(gradient clipping)} The $\ell_2$ norm of the stochastic gradient on each sample in each client is bounded: $\|{\nabla}{f_{C,m}(\mathbf{w};\textbf{z})}\|_2^2\leq{G_1}$.

\textbf{Corollary 1.} \textit{(bounded local variance)} With the gradient clipping, the variance of the stochastic gradient in each client is bounded: $\mathbb{E}[||\tilde{\nabla}{f_{C,m}(\mathbf{w}_C)}||_2^2]$, $\mathbb{E}[||\tilde{\nabla}{F_{C,m}(\mathbf{w}_C)}||_2^2]$, $\mathbb{E}[||{\nabla}{F_{C}(\mathbf{w}_C)}||_2^2]\leq{G_1}$.

\textbf{Assumption 2.} \textit{(non-convexity, L-smoothness)}  The local loss function of the arbitrary $m$-th client $f_{C,m}(\mathbf{w}_C)$ or $F_{C,m}(\mathbf{w}_C)$ is non-convex and satisfy: $\|f_{C,m}(\mathbf{u})-f_{C,m}(\mathbf{v})\|_2,\|F_{C,m}(\mathbf{u})-F_{C,m}(\mathbf{v})\|_2\leq{L}\|\mathbf{u}-\mathbf{v}\|_2$.

\textbf{Definition 1.} \textit{(weighted gradient diversity)} Let $\mathbf{w}$ collect the
local iterates $(\mathbf{w}_{C,m}^{(t)})_{m=1}^M$ at step $t$, and let
$\mathbf{q}=(q_1,\ldots,q_M)$. Following
\cite{shi2024sam,yin2018gradient}, $\lambda_1$ satisfies
\begin{equation}
\Lambda(\mathbf{w},\mathbf{q})\triangleq\frac{\sum_{m=1}^Mq_m\|{{{\nabla}{F_{C,m}(\mathbf{w}_{C,m}^{(t)})}}}\|_2^2}{\|\sum_{m=1}^M{{{q_m\nabla}{F_{C,m}(\mathbf{w}_{C,m}^{(t)})}}}\|_2^2}\leq\lambda_1
\end{equation}

\textbf{Assumption 3.} \textit{(inequality and lower-bound of the distance of the distribution)} The intermediate labels for the clients originate from their original labels which are projected by the same inverse server-side model. However, the smashed data from different clients on the server originate from the original data which experience different forward passes due to the distinctive model parameters of client-side models. Therefore, we have: in the arbitrary $t$-th local iteration, $d_{C,m}^t<d_{S,m}^t$. Meanwhile, both distances have the following lower bounds:
\begin{equation}
d_{C,m}^t>B_1>0,\quad d_{S,m}^t>B_2>0, B_1<B_2
\end{equation}

\textbf{Theorem 1.} If Corollary 1, Assumption 2 and Definition 1 hold, and the local learning rate satisfies:
\begin{equation}
-\frac{\eta_C}{2}+8{\lambda_1}E^2L^2{\eta_C}^3+2\lambda_1{\eta_C}^2L\leq0\label{eqcondition}
\end{equation}
The local iteration to find the optimal point of the subproblem 1 in the equation \ref{eq4} can converge with the following speed with $T$ denoting the number of iterations.
\begin{align}
    &\frac{1}{T}\sum_{t=0}^{T-1}\|{\nabla}{F_{C}(\bar{\mathbf{w}}_C^{(t)})}\|_2^2\leq\frac{2[F_{C}(\bar{\mathbf{w}}_C^{(0)})-F_{C}(\bar{\mathbf{w}}_C^{(*)})]}{{\eta_C}T} \notag\\
    &+\frac{1}{T}\sum_{t=0}^{T-1}\{({2+2{\eta_C}L})G_1\sum_{m=1}^M{q_m}[d_{C,m}^t]\notag\\
    &+{\eta_C{L}G_1}\sum_{m=1}^{M}q_m[d_{C,m}^t]^2\}\notag\\
    &+3G_1\eta_C^2L^2{E(E+1)}+\frac{3\eta_C{L}}{2}G_1\label{eq13}
\end{align}
Here $\bar{\mathbf{w}}_C^{(t)}=\frac{1}{K}\sum_{m\in\mathcal{A}(t)}\mathbf{w}_{C,m}^{(t)}$,
and $\bar{\mathbf{w}}_C^{(*)}$ denotes an optimal reference point of $F_C$.
Appendix A proves Theorem 1.

\textbf{Corollary 2.} Training the client-side model can achieve a convergence rate of $\mathcal{O}(\frac{1}{\sqrt{T}})$, by plugging:
\begin{equation}
\eta_C=\frac{1}{\sqrt{TE}{(2L\sum_{m=1}^Mq_mB_1}+L\sum_{m=1}^Mq_mB_1^2)}
\end{equation}
into the inequality \ref{eq13}. We can have:
\begin{align}
    &\frac{1}{T}\sum_{t=0}^{T-1}\|{\nabla}{F_{C}(\bar{\mathbf{w}}_C^{(t)})}\|_2^2\leq\notag\\
    &\frac{\tau_*(2L\sum_{m=1}^Mq_mB_1+L\sum_{m=1}^Mq_mB_1^2)}{\sqrt{T}}+{2G_1}\sum_{m=1}^Mq_m[d_{C,m}^0]\notag\\
    &+\frac{G_1}{\sqrt{TE}}+\frac{3G_1(E+1)}{T(2B_1+B_1^2)^2}+\frac{3G_1}{2\sqrt{TE}(2B_1+B_1^2)}\notag\\
    % &=\mathcal{O}(\frac{\sqrt{E}}{\sqrt{T}})+\mathcal{O}(\frac{1}{\sqrt{TE}})+\mathcal{O}(\frac{E+1}{T})\notag\\
    &\Leftrightarrow\mathcal{O}(\frac{\sqrt{E}}{\sqrt{T}})+\mathcal{O}(\frac{1}{\sqrt{TE}})\Leftrightarrow\mathcal{O}(\frac{1}{\sqrt{T}})
\end{align}
where $\tau_*=2\sqrt{E}[F_{C}(\bar{\mathbf{w}}_C^{(0)})-F_{C}(\bar{\mathbf{w}}_C^{(*)})]$.

\textbf{Corollary 3.} Following the same proof process as Theorem 1 and Corollary 2, the inverse server-side model can also achieve an $\mathcal{O}(\frac{1}{\sqrt{T}})$ convergence rate when $\eta_S=\frac{1}{\sqrt{TE}{(2L\sum_{m=1}^Mq_mB_2}+L\sum_{m=1}^Mq_mB_2^2)}$. Since $B_1<B_2$, we set $\eta_C>\eta_S$ in the practical implementation to achieve proximate convergence rate for the client- and inverse server-side models.

\textbf{Corollary 4.} To reach an $\epsilon$-level accuracy, SplitMe requires at least $K_{\epsilon}^{co}\geq\mathcal{O}(\frac{(E+1)^2}{E^2\epsilon^2})$ global communication rounds. This corollary means the transmission between xApp and rApp can be further reduced with an increased requirement on the computational resources of the non-RT-RIC and near-RT-RIC.
\section{System Optimization}
\subsection{SFL Resource Modeling}
In order to transfer the smashed data and local model parameters from xApps to their rApps, the limited uplink bandwidth resources are to be assigned. Meanwhile, the computational resources of the near-RT-RICs, utilized for the client-side model training, are scarce. So, these resources need to be judiciously allocated to determine the learning time, communication rounds and therefore impact the performance of the final SFL model. 

For the communication part, a portion of O-RAN's m-plane fiber link is budgeted for uploading with a total bandwidth $B$. Let $b_m^k$ be the bandwidth fraction allocated to the $m$-th client in the global communication round $k$. As a result, its allocated bandwidth is $b_m^kB$ and $\mathbf{b}^k=(b_1^k,\dots,b_M^k),\sum_{m=1}^{M}b_m^k=1~\forall{k}$ represents the bandwidth allocation vector.

In each round, the concerned rApp determines a fraction of xApps (clients) and their corresponding rApps to participate in the model training due to the delay constraints originating from the O-RAN control loop \cite{wg2023cloud}. We define a binary variable $a_m^k\in\{1,0\}$ to decide whether the $m$-th xApp and rApp are selected in the current round. The selection vector is $\mathbf{a}^k=(a_1^k,\dots,a_M^k)$, with $a_m^k=1$ exactly when $m\in\mathcal{A}_k$. Only when $a_m^k=1$, at least a minimum bandwidth $b_{min}$ is to be allocated to the client $m$, i.e. $b_m^k\geq{b_{min}},b_{min}\leq\frac{1}{M}$. Otherwise, there will be no bandwidth to allocate. Therefore, following \cite{singh2024communication,fu2025multi}, we formulate the communication resource usage cost in each global communication round as:
\begin{equation}
R^{co}=\sum_{m=1}^MR^{co}_m=\sum_{m=1}^Ma_m^kb_m^kBp_{c}
\end{equation}
where $p_{c}$ is the unit cost of bandwidth usage. Meanwhile, let $R_m^{cp}$ denote the total computational resource cost of the $m$-th xApp and rApp, which depends on the number of local updates and local processing power. Then, different from the previous works \cite{singh2022joint,singh2024communication}, the total computational resource cost in each round of the non-RT-RICs and near-RT-RIC should be simultaneously considered, which is:
\begin{equation}
R^{cp}=\sum_{m=1}^MR^{cp}_m=\sum_{m=1}^Ma_m^kE(Q_{C,m}+Q_{S,m})p_{tr}
\end{equation}
where $Q_{C,m}$ and $Q_{S,m}$ are the processing time to consume one batch of data on the $m$-th xApp and rApp respectively. $p_{tr}$ is the per-unit-time usage cost.

\subsection{Latency Modeling}
We consider the same synchronous training mode as \cite{lin2024efficient,xu2024accelerating}. In other words, the non-RT-RIC starts training the inverse server-side model only when it receives the intermediate feature matrix and model parameters from the participating near-RT-RICs. However, different from them, SplitMe does not require backpropagating the gradients from the non-RT-RIC to the near-RT-RICs.  Therefore, in each global training round, our SFL tasks are spanned over five operations: (i) downloading the latest client-side model and class-wise targets at participating near-RT-RICs, (ii) client-side model training, (iii) uploading smashed data and local model parameters to the non-RT-RIC, (iv) inverse server-side model training, and (v) model aggregation and broadcasting the aggregated model to all rApps. We do not consider the delay in the downlink phase and the broadcast in the bus because they are negligible compared to the uplink delay as a result of high-speed communication. Let the communication time required in uploading from the $m$-th near-RT-RIC to the non-RT-RIC be $T_m^{co}(k)$ in the uplink phase. Let $d$ and $S_{m}$ be the datasize of the entire model and intermediate feature matrix of the $m$-th near-RT-RIC. Then, the total learning time in one training round is:
\begin{equation}
T^{total}(k)=\max_{m\in\mathcal{A}_k}\{EQ_{C,m}+T_m^{co}(k)\}+\max_{m\in\mathcal{A}_k}\{EQ_{S,m}\}
\end{equation}
where $T_m^{co}(k)$ is calculated as:
\begin{equation}
    T_m^{co}(k)=\frac{S_m+\omega d}{b_m^kB},\quad m\in\mathcal{A}_k
\end{equation}
where $\omega$ is the proportion of the number of client-side model parameters to that of the entire model parameters.
\subsection{Problem Formulation}
With the modeling of resources and latency, we can formulate the total cost in each round as:
\begin{align}
    &cost(k)=\rho(R^{co}+R^{cp})+(1-\rho)T^{total}(k) \notag\\
    &=\rho{(\sum_{m=1}^Ma_m^kb_m^kBp_{c}+\sum_{m=1}^Ma_m^kE(Q_{C,m}+Q_{S,m})p_{tr})}\notag\\
    &+(1-\rho)\bigl(\max_{m\in\mathcal{A}_k}\{EQ_{C,m}+T_m^{co}(k)\}\notag\\
    &\qquad\qquad+\max_{m\in\mathcal{A}_k}\{EQ_{S,m}\}\bigr)
\end{align}
To express the total cost over $K_\epsilon$ global rounds, let
$\overline{\mathrm{cost}}$ be the average cost per round. Then
\begin{equation}
    K_\epsilon\overline{\mathrm{cost}}
    =K_\epsilon\frac{1}{K_\epsilon}\sum_{k=1}^{K_\epsilon}cost(k)
    =\sum_{k=1}^{K_\epsilon}cost(k).
\end{equation}
Therefore, this optimization problem can be reduced to optimizing the total cost in each round:
\begin{align}
    \text{\textbf{P}:}&\underset{\mathbf{a}^k,\mathbf{b}^k,E}{\arg
    \min}~K_{\epsilon}cost(k)\\
    &\text{subject to:}\sum_{m=1}^Ma_m^kb_m^kB=B,\tag{22a}\\
    &~~~~~~~~~~~~\sum_{m=1}^Mb_m^k=1,\tag{22b}\\
    &~~~~~~~~~~~~b_{min}\leq{b_m^k}\leq1;\forall{m}\in\mathcal{A}_k,\tag{22c}\\
    &~~~~~~~~~~~~a_m^k\in\{1,0\},\tag{22d}\\
    &~~~~~~~~~~~~E\in\{1,\dots,N\},\tag{22e}\\
    &~~~~~~~~~~~~K_{\epsilon}=\mathcal{O}(\frac{(E+1)^2}{E^2\epsilon^2}).\tag{22f}
\end{align}
Here $K_\epsilon$ is the communication-round count $K_\epsilon^{co}$ in Corollary~4, and $N$ is the upper bound on local updates per round.
\subsection{Solution}
The problem formulated is a non-convex problem due to the non-convex objective function and constraint (22a). Furthermore, there does not exist a closed-form solution to this problem. Thus, we take an approach of decomposition method by alternately solving two subproblems: deadline-aware local trainers selection (P1) and computational and communication resource allocation (P2). Then, we use the solution of P1 to reshape the problem of P2 and solve P2 ultimately. We solve the two problems at the beginning of each training round with $E=E_{\mathrm{initial}}$ in the first training round.

For P1, we use the proposition that a larger fraction of trainers in each round saves the total time required for a global FL model to attain the desired model performance \cite{mcmahan2017communication}. The optimization problem of P1 can be transformed to maximizing the set of selected local trainers while the number of local updates $E$ is fixed and the deadline of the O-RAN control loop is not violated:
\begin{align}
    &\underset{\mathcal{A}_k}{\max}\{\vert{\mathcal{A}_k}\vert\}\\
    &\text{s.t.}E(Q_{C,m}+Q_{S,m})+t_{\max}^k\leq t_{round},
     \quad m\in\mathcal{A}_k.\tag{23a}
\end{align}
Here $t_{\max}^{k}$ is the maximum uplink time
estimated before round $k$. At initialization, the estimate assumes
equal bandwidth across all $M$ clients:
\[
    t_{\max}^{1}=\max_{m\in\mathcal{M}}\frac{M(S_m+\omega d)}{B}.
\]
Algorithm~\ref{al2} updates the estimate after bandwidth allocation.
The latency model treats feature extraction, trainer selection, and
resource allocation as negligible relative to local updates and the
uplink. The deadline $t_{round}$ is specific to the network
slice.
\begin{algorithm2e}[!tb]
\caption{Deadline-Aware Selection of Trainers}\label{al1}
\LinesNumbered
\SetKwInOut{Input}{Input}\SetKwInOut{Output}{Output}
\Input{Candidate clients $\mathcal{M}$, local update budget $E$,
uplink time estimate $t_{\max}^k$, and deadline $t_{round}$}
\Output{Participating clients $\mathcal{A}_k$}
$\mathcal{A}_k\gets\varnothing$\;
\ForEach{$m\in\mathcal{M}$}{
    \If{$E(Q_{C,m}+Q_{S,m})+t_{\max}^k\leq t_{round}$}{
        Add $m$ to $\mathcal{A}_k$\;
    }
}
\end{algorithm2e}

After the phase of trainer selection, we obtain $\mathbf{a}^k$. Then, we should allocate the communication and computational resources to support uploading the smashed data and local model parameters, and training of client- and inverse server-side models. Different from the previous works, we take the number of local updates into consideration since the change of $E$ can make the targets of reducing the resource cost and reducing learning time conflicting. With the difference and substituting the defining expressions, the optimization problem (22) is reduced to the following mathematical form:
\begin{align}
    \text{\textbf{P2}:}&\underset{\mathbf{b}^k,E}{\arg
    \min}~K_{\epsilon}cost^*(k)\\
    &\text{subject to:} \text{(22a),(22b),(22c),(22e),(22f)}\notag
\end{align}
Here $cost^*(k)$ is the round cost with the selected clients
$\mathcal{A}_k$ fixed.
We solve this mixed integer programming non-convex problem by using Ipopt solver. Let $E_{\mathrm{last}}$ be the number of local updates used in the phase of trainer selection and $\hat{E}$ be the new one obtained by the solver. Only when $\hat{E}\leq{E_{\mathrm{last}}}$, $E=\hat{E}$ in the current training round. Otherwise, $E=E_{\mathrm{last}}$. Such a measurement escapes $E$ exceeding the permitted range, which can result in violating the deadline of the O-RAN control loop. Furthermore, we define $E_{\max}$ as the largest number of local updates within the SFL training, and theorem 1, corollary 2, corollary 3 and corollary 4 still hold by plugging $E=E_{\max}$. Finally, we present SplitMe with system optimization in Algorithm \ref{al2}.

\begin{algorithm2e}[!tb]
\caption{SplitMe with System Optimization}\label{al2}
\LinesNumbered
\SetKwInOut{Input}{Input}\SetKwInOut{Output}{Output}
\Input{Local datasets $\{\mathcal{D}_m\}_{m\in\mathcal{M}}$,
initial local update budget $E_{\mathrm{initial}}$,
round count $K_\epsilon$, and smoothing factor $\alpha$}
\Output{Inference parameters
$(\bar{\mathbf{w}}_C^{(T)},\mathbf{w}_S^*)$}
Initialize the global models and register the ordered labels\;
Set $t_1=0$ and $E=E_{\mathrm{initial}}$, and initialize $t_{\max}^1$ as in
Section~IV-D\;
\For{$k=1,\ldots,K_\epsilon$}{
    Select $\mathcal{A}_k$ with Algorithm~\ref{al1}\;
    Use P2 to allocate $\mathbf{b}^k$ and update $E$\;
    Set $t_{k+1}=t_k+E$\;
    \ForEach{xApp $m\in\mathcal{A}_k$ in parallel}{
        Download the client-side model and the class-wise targets\;
        Form $\tilde{\mathcal{D}}_m$ and update the local
        client-side model using \eqref{eq5}\;
        Upload the local parameters and smashed data $\mathbf{Z}_m^k$\;
    }
    \ForEach{associated rApp $m\in\mathcal{A}_k$ in parallel}{
        Form $\check{\mathcal{D}}_m$ from the stored labels and
        $\mathbf{Z}_m^k$, then update the inverse server-side
        model using \eqref{eq6}\;
    }
    Aggregate the client-side and inverse server-side models separately\;
    $t_{\max}^{k+1}\gets\alpha t_{\max}^k+
    (1-\alpha)\max_{m\in\mathcal{A}_k}T_m^{co}(k)$\;
}
Set $T=t_{K_\epsilon+1}$. Recompute smashed data for
$\mathcal{R}=\mathcal{A}_{K_\epsilon}$
using $\bar{\mathbf{w}}_C^{(T)}$\;
Reconstruct $\mathbf{w}_S^*$ layer by layer using \eqref{eq8}\;
Send the combined model to the SMO for deployment\;
\end{algorithm2e}
\section{Numerical Results}
\subsection{Experimental Setting}
\textbf{Testbed setup.} We use an NVIDIA GeForce RTX 4090 GPU workstation as an emulation environment. We employ 8 RTX 4090 GPUs as the cloud server (non-RT-RIC) and 50 Intel(R) Core(TM) i5-8265U CPUs as 50 clients (near-RT-RICs). Each client corresponds to one rApp, which is randomly associated with one GPU on the server. Since the cloud server usually utilizes the Gigabit Ethernet, the total bandwidth of it is set as 1 Gbps. The main settings are summarized in Table. \ref{tab3}.

\textbf{Datasets and Models.} To prove the effectiveness of SplitMe in the 5G network slicing scenario, we conduct our experiments on the COMMAG O-RAN dataset \cite{bonati2021intelligence}, which is open-source and generated through the Colosseum test platform in a 5G network simulation. It takes into account the actual base station locations within a 0.11 square kilometer area in Rome, Italy, and uses 40 user equipment (UE) to generate three types of traffic: eMBB, mMTC, and uRLLC. A ten-layer DNN is trained, just as in \cite{couto2024survey}, to solve the traffic classification problem. However, such a model is solved in a federated setting where each near-RT-RIC is fed with slice-specific network data and thus stores only one type of traffic data, resulting in data heterogeneity and making the training difficult. In terms of the model splitting, we cut $20\%$ (i.e. two) layers to the clients and the rest to the cloud server to reduce the clients' workload. 
\begin{table}[!tbp]
\renewcommand{\arraystretch}{1.0}
\setlength\tabcolsep{1.8pt}
\caption{Experimental Settings.}
\begin{center}
\centering
\begin{tabular}{c|l|l}
%添加顶部横线 
\Xhline{1.5 pt}
%输入标题
% \cite{2-3}
% \cline{2-7}
% \Xhline{1.5 pt}
%添加标题和内容之间的横线
% \Xhline{0.5 pt}
\centering
% \cline{2-4}

\textbf{Parameter}&\textbf{Description}&\textbf{Value}\\
\Xhline{0.5 pt}
$M$&Maximum Number of Local Trainers&50\\
$B$&Total Bandwidth Budget for SFL Training&1Gbps\\
$Q_{C,m}$&Per Batch Processing time of the $m^{th}$ xApp&${U(0.34,0.46)}$ms\\
$Q_{S,m}$&Per Batch Processing time of the $m^{th}$ rApp&${U(1.2,1.6)}$ms\\
$p_{c}$&Per Unit Cost Communication Cost&1\\
$p_{tr}$&Per Unit Computation Cost&1\\
$b_{min}$&Minimum Bandwidth Allocation&1/50\\
$\omega$&Split Proportion&1/5\\
$\rho$&Pareto Trade-off Parameter&0.8\\
$t_{round}$&Slice-Specific Deadline&$U(50,100)$ms\\
$\alpha$&Heuristic Factor&0.7\\
% $\textbf{z}_{C,m}^*$&the label of one data sample on the $m$-th client, which is\\& generated from the optimal inverse server-side model\\
% $p_{C,m}^t(\textbf{z})$& the probability distribution of $\textbf{z}_{C,m}^t$, which is time-varying\\
% $p_{C,m}^*(\textbf{z})$& the probability distribution of $\textbf{z}_{C,m}^*$\\
% $d_{C,m}^t$&$\int\|{p_{C,m}^t(\textbf{z})-p_{C,m}^*(\textbf{z})}\|d\textbf{z}$\\
% ${\nabla}{f_{C,m}(\textbf{w};\textbf{z})}$&the gradient from one data sample on the $m$-th client\\
% $q_m$&the probability of choosing the $m$-th client,$\sum_{m=1}^M{q_m}=1$\\
% % $f_C(\textbf{w}_{C})$&the global loss function of the client-side model\\
% % $F_C(\textbf{w}_{C})$&the global loss function of the client-side model when the\\& inverse server-side model is optimal\\
% $\tilde{\nabla}{F_{C,m}(\textbf{w}_{C})}$&the gradient computed on a data batch $\xi_m^t\subset\tilde{\mathcal{D}}_m$ using\\& the local loss function $F_{C,m}(\textbf{w}_{C})$\\
% % ${\nabla}{F_{C,m}(\textbf{w}_C)}$&$\mathbb{E}[\tilde{\nabla}{F_{C,m}(\textbf{w}_C)}]$\\
% $\nabla{F_C(\textbf{w}_{C})}$&${\frac{1}{K}\sum_{m\in\mathcal{A}_t}({\nabla}{F_{C,m}(\textbf{w}_{C})}}=\mathbb{E}[\tilde{\nabla}{F_{C,m}(\textbf{w}_C)}])$\\
\Xhline{1.5 pt}
\end{tabular}
\label{tab3}
\end{center}
\vspace{-0.6cm}
\end{table}

\textbf{Metrics and baselines.}
For a comprehensive comparison of the performance on the O-RAN system, we consider the number of selected trainers, total communicated volume (MB), total training time (s) and communication resource cost. We compare SplitMe with the following three classical baselines to prove its superiority and originality in the O-RAN system. All the hyperparameter settings below are determined in order not to violate the slice-specific deadline while achieving the best convergence performance.

\textit{1) Federated Averaging (FedAvg) \cite{mcmahan2017communication}:} FedAvg with fixed number of selected clients $K=10$ and local updates $E=10$ is considered. This serves as the basic FL framework without model splitting and system optimization on resources.

\textit{2) Split Federated Learning (SFL) \cite{thapa2022splitfed}:} The vanilla SFL framework with fixed number of selected clients $K=20$ and local updates $E=14$ is considered. This serves as the basic SFL framework without any variation for O-RAN.

\textit{3) O-RAN Federated Learning (O-RANFed) \cite{singh2022joint}:} This is an FL framework that uses a deadline-aware trainer selection and bandwidth allocation for the O-RAN system. This serves as an FL variant with system optimization for O-RAN but without model splitting.

\subsection{Performance Evaluation}

Since SplitMe requires 30 training rounds to achieve the highest accuracy (i.e. $83\%$), which overwhelms other baselines, we merely record its metrics within 30 rounds. 

\begin{figure}[!tbp]
\centering
\subfloat[Selected Client Number]{\includegraphics[width=4.4cm]{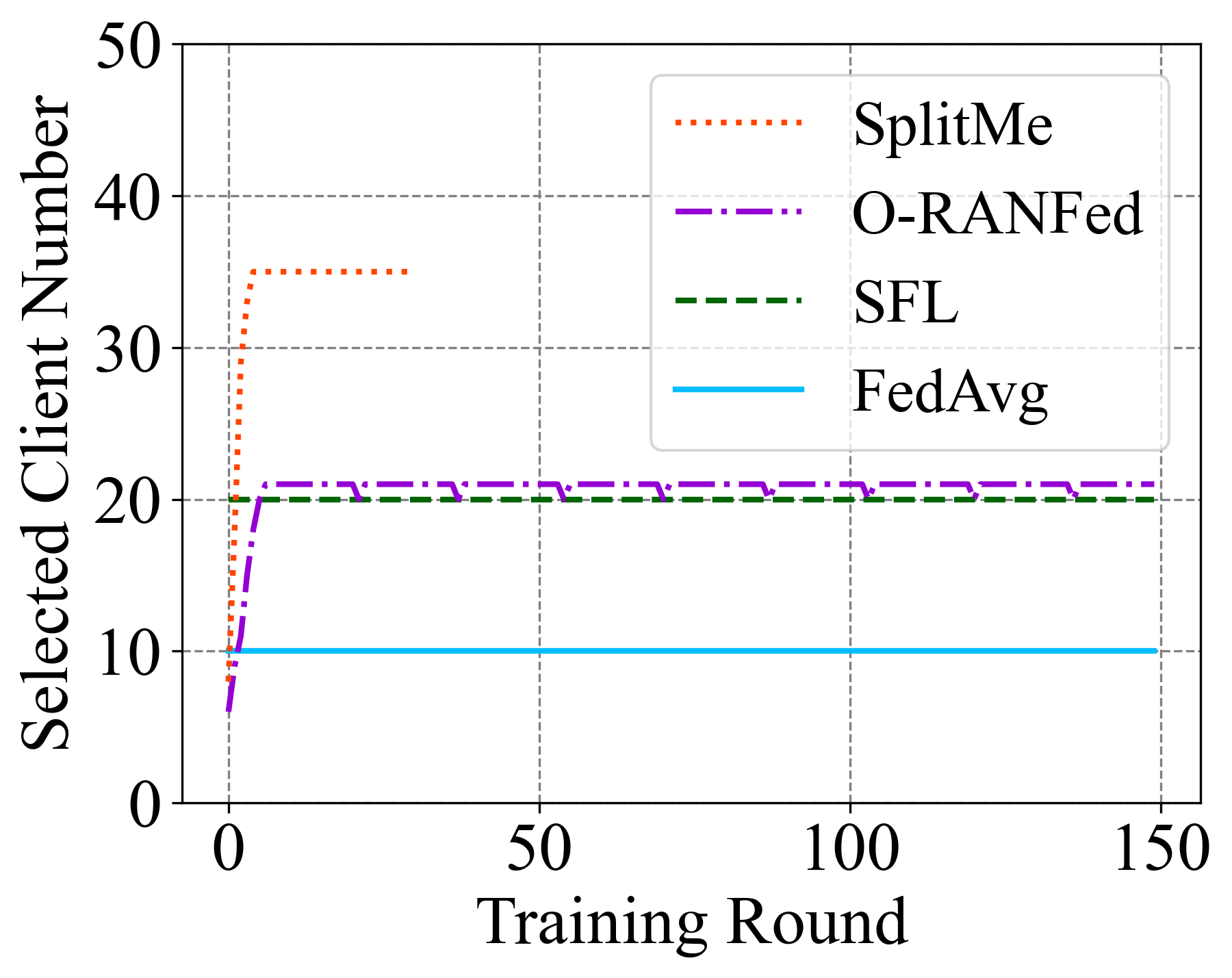}\label{ex1_1}%
}
% \hfil
\subfloat[Communication Volume]{\includegraphics[width=4.4cm]{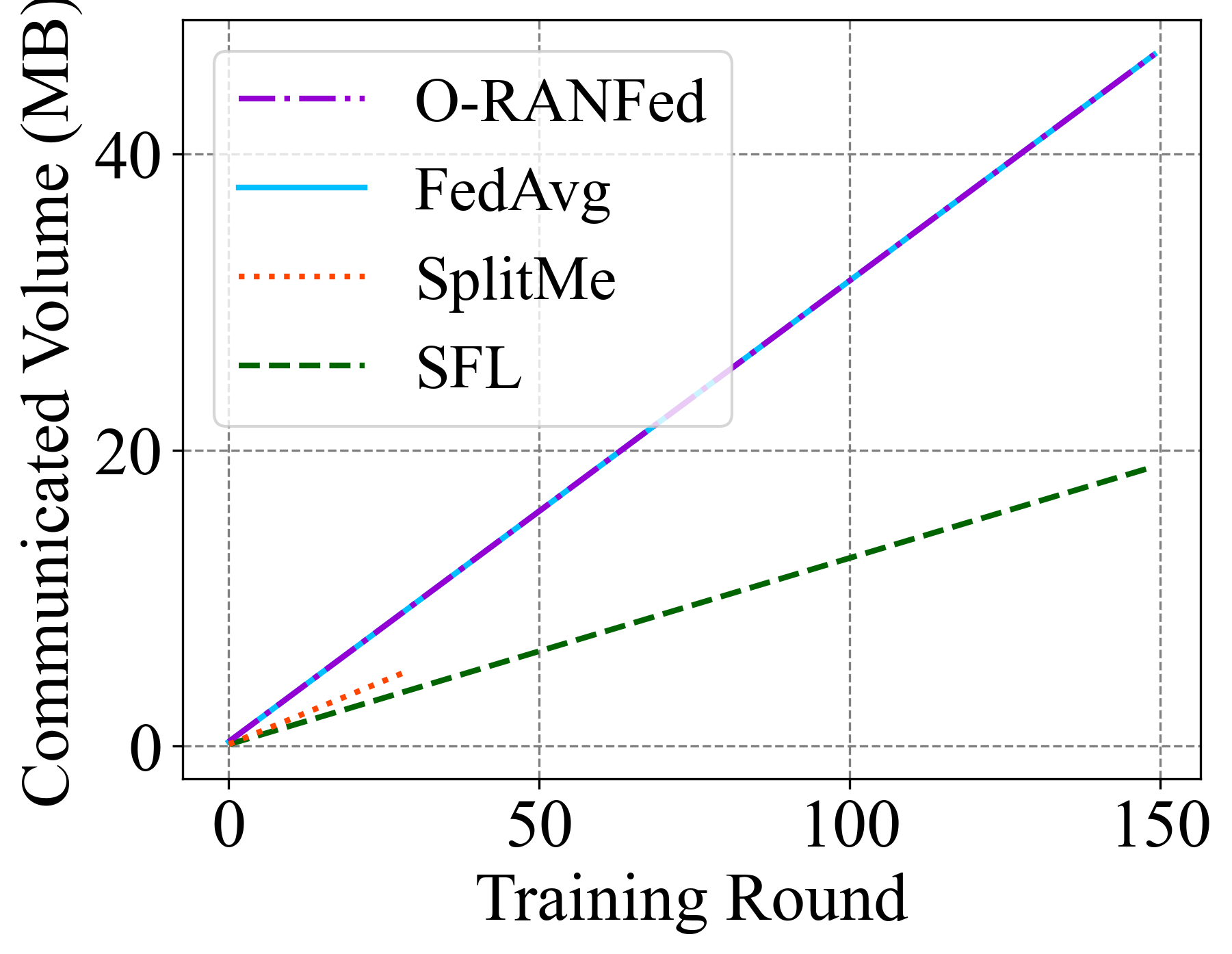}%
}
\caption{The number of selected clients and accumulated communication volume within 150 training rounds. Specially, SplitMe requires only 30 rounds to complete training.}
\label{ex1}
\vspace{-0.3cm}
\end{figure}

\textit{Number of selected trainers per round:} Fig. \ref{ex1_1} shows the number of selected trainers (clients) of each framework and SplitMe incorporates the most trainers, whose number is up to 35. Although SFL demonstrates that a lower local computation time on the near-RT-RIC can lead to more trainers to be selected, SplitMe further enlarges this advantage. That is because our deadline-aware selection algorithm and resource allocation algorithm start from an extreme point ($E_{\mathrm{initial}}=20$, $\lvert\mathcal{A}_1\rvert=8$). In each training round, the selection algorithm selects more trainers as possible while the resource allocation algorithm adaptively reduces the number of local updates to solve \textbf{P2}. Therefore, SplitMe can incorporate a larger fraction of trainers, facilitating the convergence.

  \begin{figure}[!tbp]
\centering
\subfloat[Test Accuracy]{\includegraphics[width=4.4cm]{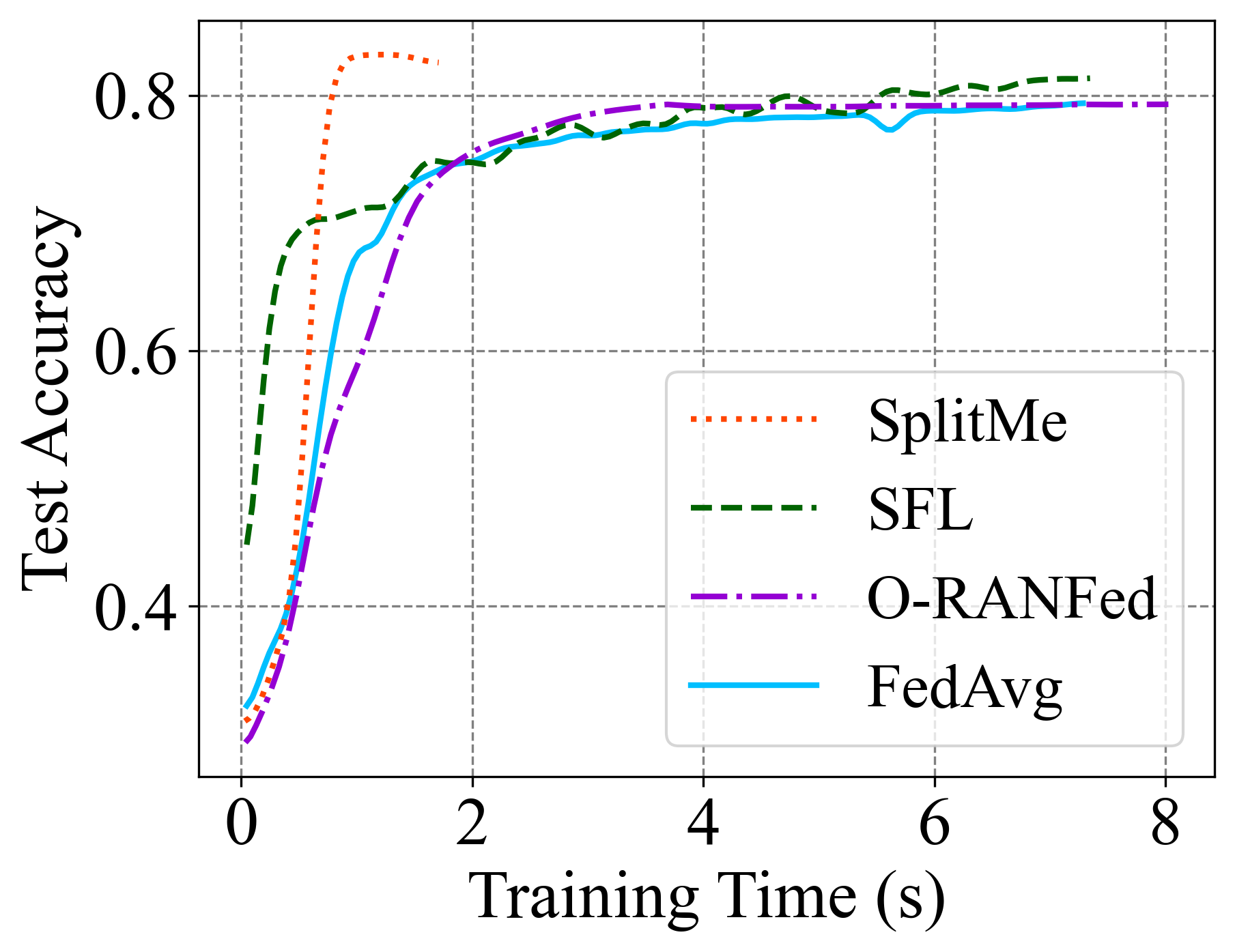}\label{ex2_1}%
} 
% \hfil
\subfloat[Communication Cost]{\includegraphics[width=4.4cm]{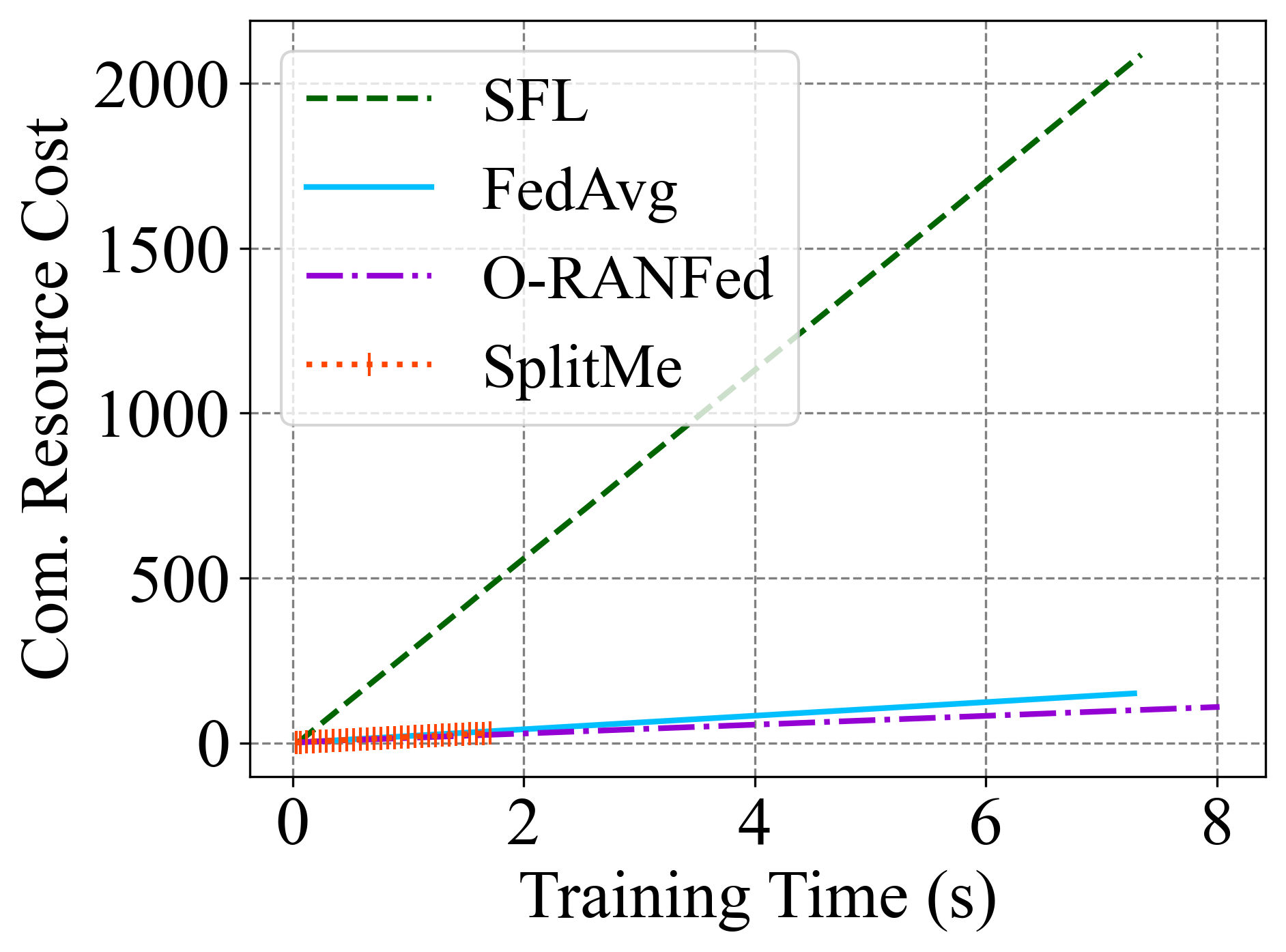}%
}
\caption{The test accuracy and communication resource cost versus total training time curve. SplitMe spends 30 rounds to reach the highest accuracy and thus requires the least training time.}
\label{ex2}
\vspace{-0.3cm}
\end{figure}

\textit{Training time:} Fig. \ref{ex2_1} shows that our SplitMe achieves $3\%$ increase on the final accuracy and requires significantly smaller training time, compared with the baselines. Due to the larger fraction of trainers and the model splitting technique, the convergence is enhanced and the computation burden on near-RT-RIC is alleviated. Furthermore, the transmission of the intermediate feature matrix and the split model is much smaller than transferring the entire model. The time cost of communication is thus reduced and moreover optimized by the bandwidth allocation in solving \textbf{P2}. As a result, with the enhanced convergence, lower client-side computation time and high-performance non-RT-RIC, SplitMe requires extremely few training rounds and thus consumes a shorter training time.
  \begin{figure}[!tbp]
\centering
\subfloat[Cifar-10]{\includegraphics[width=4.4cm]{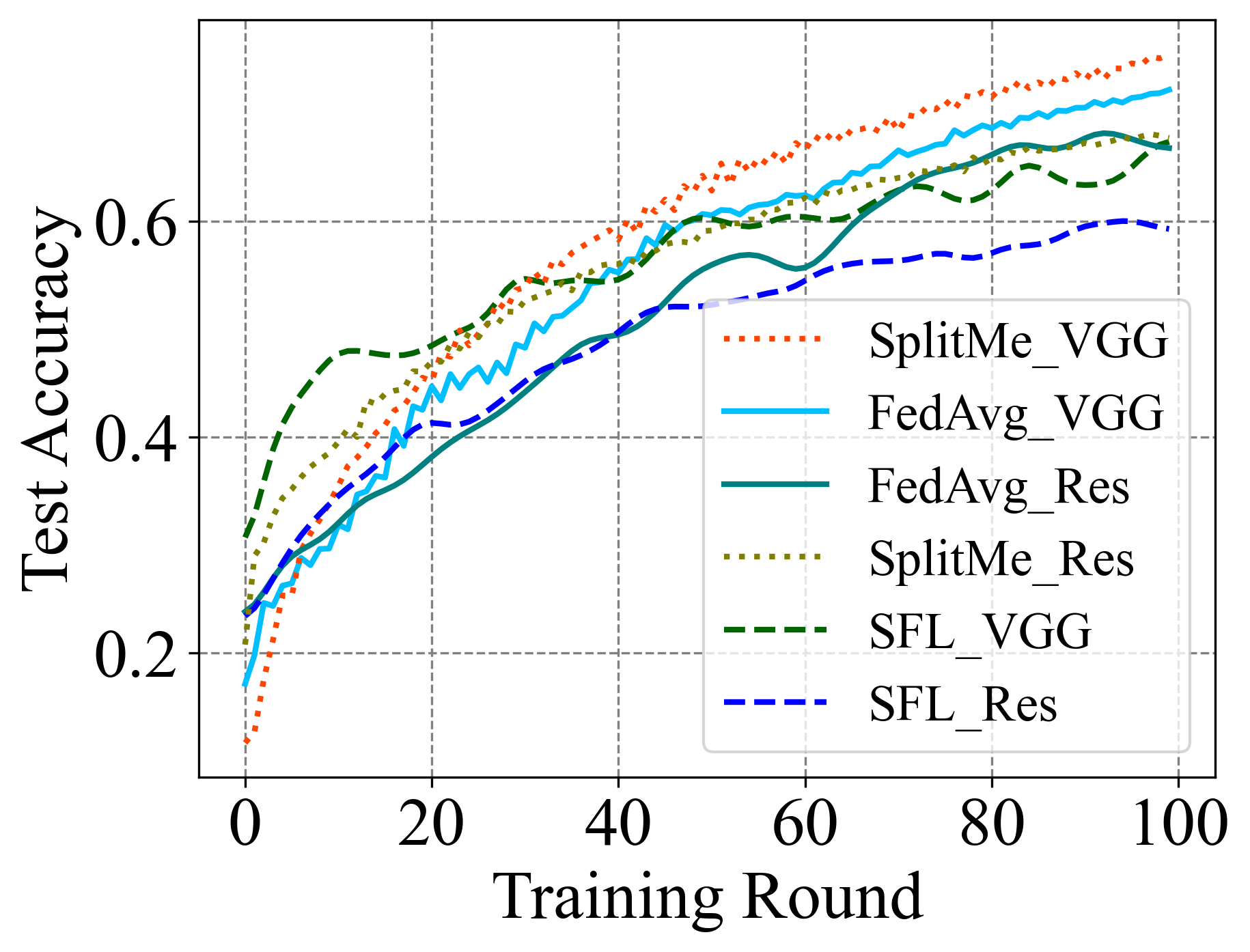}%
} 
% \hfil
\subfloat[Cifar-100]{\includegraphics[width=4.4cm]{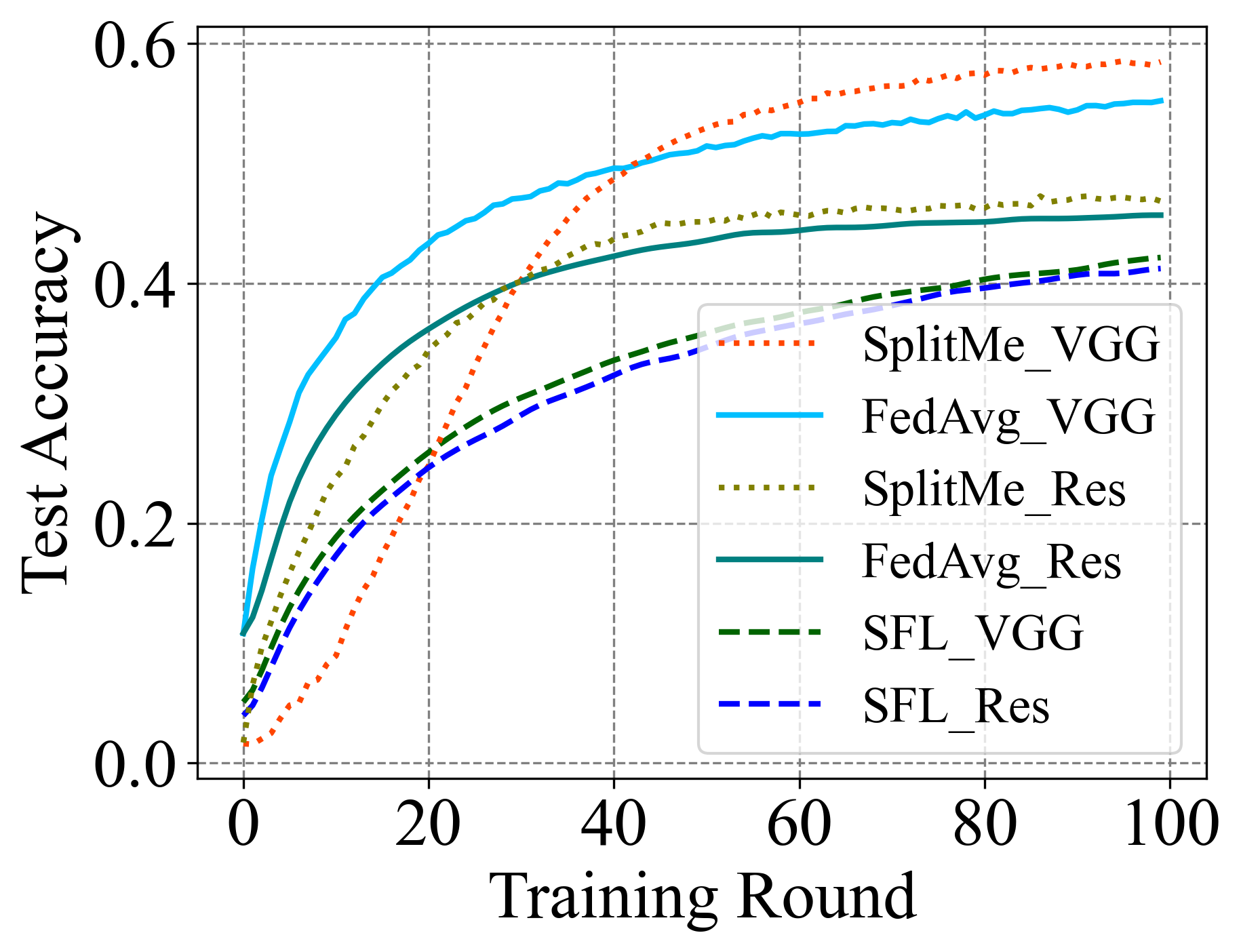}%
}
\caption{Extra experiment on the Cifar-10/100 datasets \cite{krizhevsky2009learning}. We train VGG-11 \cite{simonyan2015very} and ResNet-18 models \cite{he2016deep} on these datasets to evaluate the generality of SplitMe on computer vision tasks.}
\vspace{-0.3cm}
\end{figure}

\textit{Communication volume:} Due to the model splitting, communicating the intermediate feature matrix and the split model to the rApps is more efficient than communicating the entire model. However, the communicated volume of SplitMe per round is kind of more than SFL. That is because SplitMe should input all the local data on the near-RT-RIC to the client-side model to produce the smashed data that supervise the inverse server-side model after the client-side updates. Meanwhile, SFL transfers the intermediate feature matrix in each local update of the current training round to the server. The number of local updates set in SFL does not cover all data and thus SFL requires a bit lower communication volume per training round. But most importantly, the total communication volume of SplitMe is greatly smaller than other baselines, thanks to the fewest training rounds (i.e. the communication rounds).

\textit{Communication Resource Cost:} Due to our proposed mutual learning method, the inverse server-side model starts its updates after the client-side model finishes its local updates. The frequent communication between the client and the server in SFL is thus eliminated in SplitMe. As a result, our SplitMe can achieve the same communication frequency as FL and O-RANFed while its total cost is still the state-of-the-art (SOTA) thanks to the fewest training rounds.

\section{Conclusions}
In this paper, we have proposed SplitMe, an SFL framework that independently trains the client-side model and the inverse server-side model to eliminate the unaffordable communication resource cost for deploying SFL in the O-RAN system. Layer-wise analytic reconstruction recovers the forward server-side model for inference. Furthermore, a system modeling of a joint optimization problem to minimize the resource cost and learning time for the O-RAN system has been proposed. A deadline-aware local trainer selection algorithm and resource allocation algorithm with adaptive local updates have been designed to solve the subproblems of it respectively. Our numerical results on the COMMAG O-RAN dataset indicate that SplitMe can swiftly achieve the SOTA accuracy ($83\%$) with an $8\times$ speedup. Meanwhile, its communication volume and resource cost prove that we successfully reduced the multiple-communication-one-round level of SFL to the one-communication-one-round level.
% To make your 
% equations more compact, you may use the solidus (~/~), the exp function, or 
% appropriate exponents. Italicize Roman symbols for quantities and variables, 
% but not Greek symbols. Use a long dash rather than a hyphen for a minus 
% sign. Punctuate equations with commas or periods when they are part of a 
% sentence, as in:
% \begin{equation}
% a+b=\gamma\label{eq}
% \end{equation}

% Be sure that the 
% symbols in your equation have been defined before or immediately following 
% the equation. Use ``\eqref{eq}'', not ``Eq.~\eqref{eq}'' or ``equation \eqref{eq}'', except at 
% the beginning of a sentence: ``Equation \eqref{eq} is . . .''

\section*{Appendix A \\ Proof of Theorem 1}
Due to the L-smoothness of the local loss function, we have:
\begin{equation}
\begin{aligned}[b]
&\frac{1}{T}\sum_{t=0}^{T-1}\mathbb{E}_{\mathcal{A}(t)}
  \bigl[\mathbb{E}[F_C(\bar{\mathbf{w}}_C^{(t+1)})
    -F_C(\bar{\mathbf{w}}_C^{(t)})]\bigr]\\
&\leq \frac{1}{T}\sum_{t=0}^{T-1}
  \underbrace{-\eta_C\mathbb{E}_{\mathcal{A}(t)}\biggl[
    \mathbb{E}\biggl[\biggl\langle
      \nabla F_C(\bar{\mathbf{w}}_C^{(t)}),
      \frac{1}{K}\sum_{m\in\mathcal{A}(t)}
        \tilde{\nabla}f_{C,m}(\mathbf{w}_{C,m}^{(t)})
    \biggr\rangle\biggr]\biggr]}_{C_{2,t}}\\
&\quad+\frac{\eta_C^2L}{2}\frac{1}{T}\sum_{t=0}^{T-1}
  \mathbb{E}_{\mathcal{A}(t)}\biggl[
    \underbrace{\mathbb{E}\biggl[
      \biggl\|\frac{1}{K}\sum_{m\in\mathcal{A}(t)}
        \tilde{\nabla}f_{C,m}(\mathbf{w}_{C,m}^{(t)})\biggr\|_2^2
    \biggr]}_{C_{1,t}}\biggr].
\end{aligned}
\label{eqstart}
\end{equation}
\begingroup
\allowdisplaybreaks[2]
For a fixed local step $t$, we first bound $C_{1,t}$:
\begin{align}
&C_{1,t}\underset{(a)}{\leq}\mathbb{E}[\|{\frac{1}{K}\sum_{m\in\mathcal{A}(t)}(\tilde{\nabla}{f_{C,m}(\mathbf{w}_{C,m}^{(t)})}-\tilde{\nabla}{F_{C,m}(\mathbf{w}_{C,m}^{(t)})}})\|_2^2]\notag\\
&+\mathbb{E}[\|{\frac{1}{K}\sum_{m\in\mathcal{A}(t)}(\tilde{\nabla}{F_{C,m}(\mathbf{w}_{C,m}^{(t)})}}-{\nabla}{F_{C,m}(\mathbf{w}_{C,m}^{(t)})})\|_2^2]\notag\\
&+\|{{\frac{1}{K}\sum_{m\in\mathcal{A}(t)}{\nabla}{F_{C,m}(\mathbf{w}_{C,m}^{(t)})}}}\|_2^2+\notag\\
&2\sqrt{G_1}\mathbb{E}[\|{\frac{1}{K}\sum_{m\in\mathcal{A}(t)}(\tilde{\nabla}{f_{C,m}(\mathbf{w}_{C,m}^{(t)})}-\tilde{\nabla}{F_{C,m}(\mathbf{w}_{C,m}^{(t)})}})\|_2]\notag\\
&\underset{(b)}{\leq}\frac{1}{K}\sum_{m\in\mathcal{A}(t)}[\int\|{\nabla{f_{C,m}(\mathbf{w}_{C,m}^{(t)};\textbf{z})}}\|\|{p_{C,m}^t(\textbf{z})-p_{C,m}^*(\textbf{z})}\|d\textbf{z}]^2\notag\\
&+{{\frac{4}{K}\sum_{m\in\mathcal{A}(t)}}\|{{{\nabla}{F_{C,m}(\mathbf{w}_{C,m}^{(t)})}}}\|_2^2}+\frac{3}{2}G_1+\notag\\
&\sqrt{G_1}\frac{2}{K}\sum_{m\in\mathcal{A}(t)}[\int\|{\nabla{f_{C,m}(\mathbf{w}_{C,m}^{(t)};\textbf{z})}}\|\|{p_{C,m}^t(\textbf{z})-p_{C,m}^*(\textbf{z})}\|d\textbf{z}]\notag\\
&\leq{\frac{3G_1}{K}}\sum_{m\in\mathcal{A}(t)}[d_{C,m}^t]^2+{{\frac{4}{K}\sum_{m\in\mathcal{A}(t)}}\|{{{\nabla}{F_{C,m}(\mathbf{w}_{C,m}^{(t)})}}}\|_2^2}+\frac{3}{2}G_1
\end{align}
where $\text{(a)}$ comes from $\mathbb{E}[x^2]=\mathbb{E}[[x-\mathbb{E}[x]]^2]+\mathbb{E}[x]^2$ and the Cauchy--Schwarz inequality. (b) comes from Jensen's inequality. Then, following \cite{haddadpour2019convergence}, we bound $C_{2,t}$:
\begin{align}
&\frac{1}{T}\sum_{t=0}^{T-1}C_{2,t}\leq\eta_C{G_1}\frac{1}{T}\sum_{t=0}^{T-1}\sum_{m=1}^M{q_m}[d_{C,m}^t]\notag\\
&-\frac{\eta_C}{T}\sum_{t=0}^{T-1}
    \mathbb{E}_{\mathcal{A}(t)}\biggl[\mathbb{E}\biggl[\biggl\langle
    \nabla F_C(\bar{\mathbf{w}}_C^{(t)}),\notag\\
&\qquad \frac{1}{K}\sum_{m\in\mathcal{A}(t)}
    \tilde{\nabla}F_{C,m}(\mathbf{w}_{C,m}^{(t)})
    \biggr\rangle\biggr]\biggr]\notag\\
&\leq\frac{1}{T}\sum_{t=0}^{T-1}\{-\frac{\eta_C}{2}\|{\nabla}{F_{C}(\bar{\mathbf{w}}_C^{(t)})}\|_2^2\notag\\
&-\frac{\eta_C}{2}\|{{{\sum_{m=1}^Mq_m\nabla}{F_{C,m}(\mathbf{w}_{C,m}^{(t)})}}}\|_2^2+\eta_C{G_1}\sum_{m=1}^M{q_m}[d_{C,m}^t]\}\notag\\
&+8{\lambda_1}E^2L^2\eta_C^3\frac{1}{T}\sum_{t=0}^{T-1}[\|\sum_{m=1}^Mq_m{{\nabla}{F_{C,m}(\mathbf{w}_{C,m}^{(t)})}}\|_2^2]\notag\\
&+3G_1\eta_C^3L^2\frac{E(E+1)}{2}+\frac{\eta_C^2L}{2}\frac{1}{T}\sum_{t=0}^{T-1}[{G_1\sum_{m=1}^{M}q_m[d_{C,m}^t]^2}+\notag\\
&4\lambda_1\|{{{\sum_{m=1}^Mq_m\nabla}{F_{C,m}(\mathbf{w}_{C,m}^{(t)})}}}\|_2^2+\frac{3}{2}G_1+2{G_1}\sum_{m=1}^M{q_m}[d_{C,m}^t]]
\end{align}
Finally, we can prove Theorem 1 by plugging the bounds of $C_{1,t}$ and $\frac{1}{T}\sum_{t=0}^{T-1}C_{2,t}$ into the inequality \ref{eqstart} when the condition of \ref{eqcondition} holds.
\endgroup
% \section*{References}

% Please number citations consecutively within brackets \cite{b1}. The 
% sentence punctuation follows the bracket \cite{b2}. Refer simply to the reference 
% number, as in \cite{b3}---do not use ``Ref. \cite{b3}'' or ``reference \cite{b3}'' except at 
% the beginning of a sentence: ``Reference \cite{b3} was the first $\ldots$''

% Number footnotes separately in superscripts. Place the actual footnote at 
% the bottom of the column in which it was cited. Do not put footnotes in the 
% abstract or reference list. Use letters for table footnotes.

% Unless there are six authors or more give all authors' names; do not use 
% ``et al.''. Papers that have not been published, even if they have been 
% submitted for publication, should be cited as ``unpublished'' \cite{b4}. Papers 
% that have been accepted for publication should be cited as ``in press'' \cite{b5}. 
% Capitalize only the first word in a paper title, except for proper nouns and 
% element symbols.

% For papers published in translation journals, please give the English 
% citation first, followed by the original foreign-language citation \cite{b6}.

% \vspace{12pt}
% \color{red}
% IEEE conference templates contain guidance text for composing and formatting conference papers. Please ensure that all template text is removed from your conference paper prior to submission to the conference. Failure to remove the template text from your paper may result in your paper not being published.

\end{document}